\pdfoutput=1
\documentclass[11pt]{article}
\usepackage{coling2020}
\usepackage{times}
\usepackage{url}
\usepackage{latexsym}
\usepackage{float}
\usepackage{booktabs}

\colingfinalcopy 

\usepackage{amssymb}
\usepackage{amsmath}
\usepackage{bm}
\usepackage{tabularx}
\usepackage{comment}
\usepackage{graphicx}
\usepackage{subcaption}

\newcolumntype{L}[1]{>{\hsize=#1\hsize\raggedright\arraybackslash}X}%
\newcolumntype{R}[1]{>{\hsize=#1\hsize\raggedleft\arraybackslash}X}%
\newcolumntype{C}[1]{>{\hsize=#1\hsize\centering\arraybackslash}X}%

\newsavebox{\measurebox}

\newcommand{\mname}{ADVIN} 

\title{Automatic Discovery of Novel Intents \& Domains from Text Utterances}

\author{Nikhita Vedula \\
  Ohio State University \\
  {\tt vedula.5@osu.edu} \\\And
  Rahul Gupta, Aman Alok, Mukund Sridhar \\
  Amazon Alexa \\
  {\tt \{gupra,alokaman,harakere\}@amazon.com} 
  \\}

\begin{document}

\maketitle

\begin{abstract}

One of the primary tasks in Natural Language Understanding (NLU) is to recognize the intents as well as domains of users' spoken and written language utterances. Most existing research formulates this as a supervised classification problem with a closed-world assumption, i.e. the domains or intents to be identified are pre-defined or known beforehand. Real-world applications however increasingly encounter dynamic, rapidly evolving environments with newly emerging intents and domains, about which no information is known during model training. We propose a novel framework, ADVIN, to automatically discover novel domains and intents from large volumes of unlabeled data. We first employ an open classification model to identify all utterances potentially consisting of a novel intent. Next, we build a knowledge transfer component with a pairwise margin loss function. It learns discriminative deep features to group together utterances and discover multiple latent intent categories within them in an unsupervised manner. We finally hierarchically link mutually related intents into domains, forming an intent-domain taxonomy. ADVIN significantly outperforms baselines on three benchmark datasets, and real user utterances from a commercial voice-powered agent. 

\end{abstract}

\section{Introduction}

Numerous everyday gadgets like mobile phones or smart speaker devices consist of an NLU component to automatically understand the semantics 
of user requests. 
Comprehending the \textit{intent} and/or \textit{domain} (groups of mutually related intents) of users' natural language utterances is a key task in these devices. Deep learning models have been proposed in the literature for intent detection in diversely expressed utterances, to avoid additional feature engineering~\cite{bhargava2013,kim2016intent,liu2016attention,sun2016weakly,zhang2016joint,kim2017onenet,goo2018slot}. 
But most models are supervised or semi-supervised, i.e. they require sufficient labeled data, and can only handle a fixed number of intents and domains \textit{seen} during model training.
However, user interactions with voice-powered agents generate large amounts of unlabeled conversational text. This data often contains newly emergent, \textit{unseen} domains and intents not encountered by the learning models before. 
Expanding the capabilities of voice assistants to account for these new intents and domains would require expensive and time-consuming human labeling efforts each time a new skill, functionality or intent is to be added. Thus, it is crucial to be able to learn effectively from unlabeled text.

Zero shot techniques~\cite{kumar2017zero,xia2018} recognize intents for which no labeled training data is available. However, they require the number of the new intent types, and some prior knowledge about the new intents to be discovered to be available at test time. Such information may be unfeasible to obtain. 
Efforts have been made to break the closed-world assumption in computer vision~\cite{scheirer2014probability,bendale2015towards,bendale2016towards,fei2016breaking} as well as in the NLU literature~\cite{shu2017doc,kim2018joint,vedula2019towards,lin2019deep}. This is the paradigm of \textit{open world learning} or \textit{open set recognition}, that identifies instances with labels unseen during training. 
However, the task of discovering the actual latent categories in the instances identified as containing unseen labels is relatively under explored. \cite{shu2018unseen} attempted this for image classification, but their model could not always outperform baselines. \cite{vedula2019towards} developed a sequence tagging approach to discover unseen intents in dialog. However, unlike our proposed framework, their method (i) is restricted to intents containing an action or activity to be performed; (ii) is customized for longer utterances containing additional background context such as customer support conversations; and (iii) does not recognize an intent expressed in different ways semantically as the same intent, or group related novel intents into novel domains. 
In this work, we attempt to bridge the gap between detecting utterances belonging to new intents/domains and actually discovering a taxonomy of those new intents/domains. Though we address the problem of novel user intent and domain discovery, our technique can easily be applied to any open classification setting. 

We propose a novel three-stage framework called \mbox{\textit{\mname{}}} (\underline{A}utomatic \underline{D}iscovery of no\underline{V}el \underline{I}ntents and Domai\underline{N}s). It automatically discovers user intents and domains in massive, unlabeled text corpora, \textit{without} any prior knowledge about the intents or domains that the text may comprise of. 
Similar to an open classification setting, our method first leverages a multi-layer transformer network, BERT~\cite{devlin2018bert}, to determine if the text input is likely to contain a novel, unseen intent or not. In the next step, \mname{} discovers the latent intent categories in the above identified input utterances in an unsupervised manner, via a knowledge transfer component. 
Finally, \mname{} hierarchically links semantically related groups of the newly discovered intents to form new domains. 
\mname{} is a generic algorithm and we empirically demonstrate that it is equally effective across several task domains and fields. 
To summarize:

(i) We propose a novel, fully automated method, \mname{}, that jointly discovers newly emerging intents as well as domains in unlabeled user utterances.

(ii) Unlike existing literature, \mname{} can generalize to diverse, open-world scenarios, and is independent of the intents and domains that it has been trained upon.

(iii) We extensively evaluate \mname{} on public benchmark datasets and real-world data from a commercial voice agent, and significantly outperform baselines across various empirical configurations.

\section{Related Work} 
\label{sec:background}

Intent detection has been successfully performed in the literature via machine learning based approaches~\cite{tur2011sentence,jeong2008triangular,kim2016intent,ravuri2015comparative}, as well as deep learning models~\cite{sun2016weakly,xia2018,shivakumar2019spoken,castellucci2019multi,lin2019deep}. 
In recent years, intent detection has been jointly done with slot filling to improve its performance~\cite{xu2013convolutional,bhargava2013,liu2016attention,zhang2016joint,kim2017onenet,goo2018slot,wang2018bi}. However, the above approaches either require sufficient labeled data for each domain and intent, or some prior knowledge about the new intents to be discovered. \mname{} seeks to eliminate these restrictions.


Methods were proposed to recognize input texts belonging to novel domains~\cite{kim2018joint} and novel intents~\cite{lin2019deep,vedula2019towards}. However, unlike \mname{}, they do not jointly detect both novel domains and novel intents, cannot detect the number of novel domains and intents, and do not discover a taxonomy of the potential domain and intent types within user utterances. 
Deep clustering networks~\cite{hsu2015neural} and Bayesian non-parametric models~\cite{dundar2012bayesian,akova2012self} were proposed to identify mixture components or clusters in data. But unlike \mname{}, the clusters identified cannot automatically be mapped to unique unseen categories. 
Our work 
is also different from semi-supervised clustering techniques~\cite{fu2016semi,akova2012self,yi2015efficient,yi2013semi,hsu2015neural}, which lever small amounts of labeled data and require the classes to be discovered to be known beforehand. Having said that, we show the gains achieved by a semi-supervised version of \mname{}, in case prior knowledge is available about the novel intents (Table~\ref{tab:supervision}).
\section{Our Framework \mname{}}

Formally, we are given a corpus of training utterances $\mathcal{D_T}$ labeled with $S$ seen intents, and a corpus of unlabeled utterances $\mathcal{D_C}$ consisting of $U$ novel intents such that $S \cap U = \varnothing$. We propose a three-stage framework \mbox{\mname{}} that (i) classifies the incoming test utterance $x \in \mathcal{D_C}$ with one of the $S = (s_1, ..., s_S)$ seen intent labels or as an \textit{unseen} (\textit{novel}) intent; (ii) for utterances $\mathcal{D_X}$ predicted as having a novel intent, it discovers the latent intents $U = (u_1, ..., u_U)$ present in them; and (iii) links together mutually related novel intents discovered to form novel domains, and an intent-domain taxonomy. 

\subsection{Stage I: Detecting Instances with Novel Intents}
\label{sec:stage1}

We construct a two-step system to detect instances with \textit{novel} intents, inspired by the success of models pre-trained on large unlabeled corpora, like GPT~\cite{radford2018improving} and BERT~\cite{devlin2018bert}. 

\begin{figure}
\begin{subfigure}{.45\textwidth}
  \centering
  \includegraphics[height=4cm]{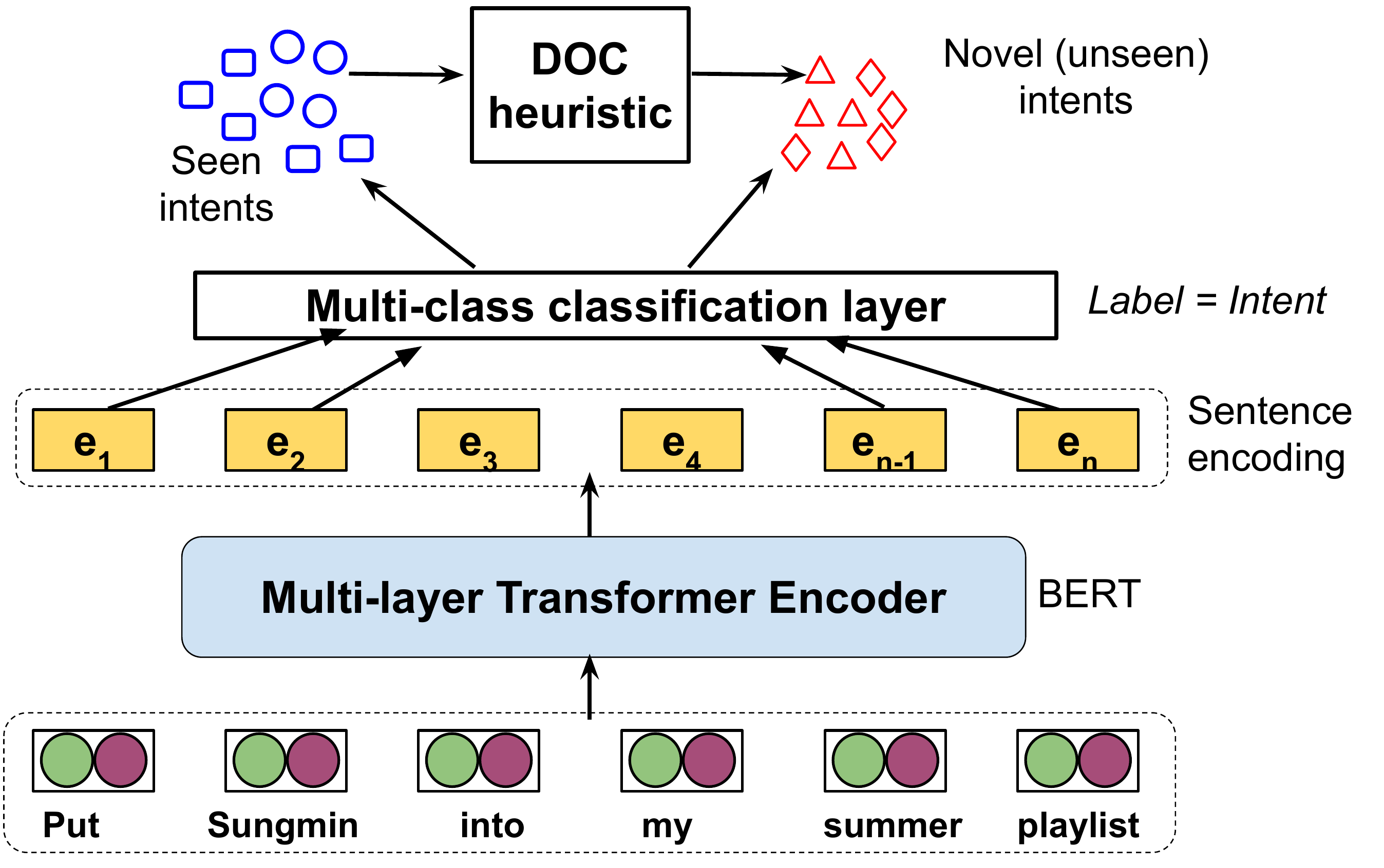}
  \caption{Detecting instances containing novel intents}
  \label{fig:stage1}
\end{subfigure}%
\begin{subfigure}{.45\textwidth}
  \centering
  \includegraphics[height=4cm]{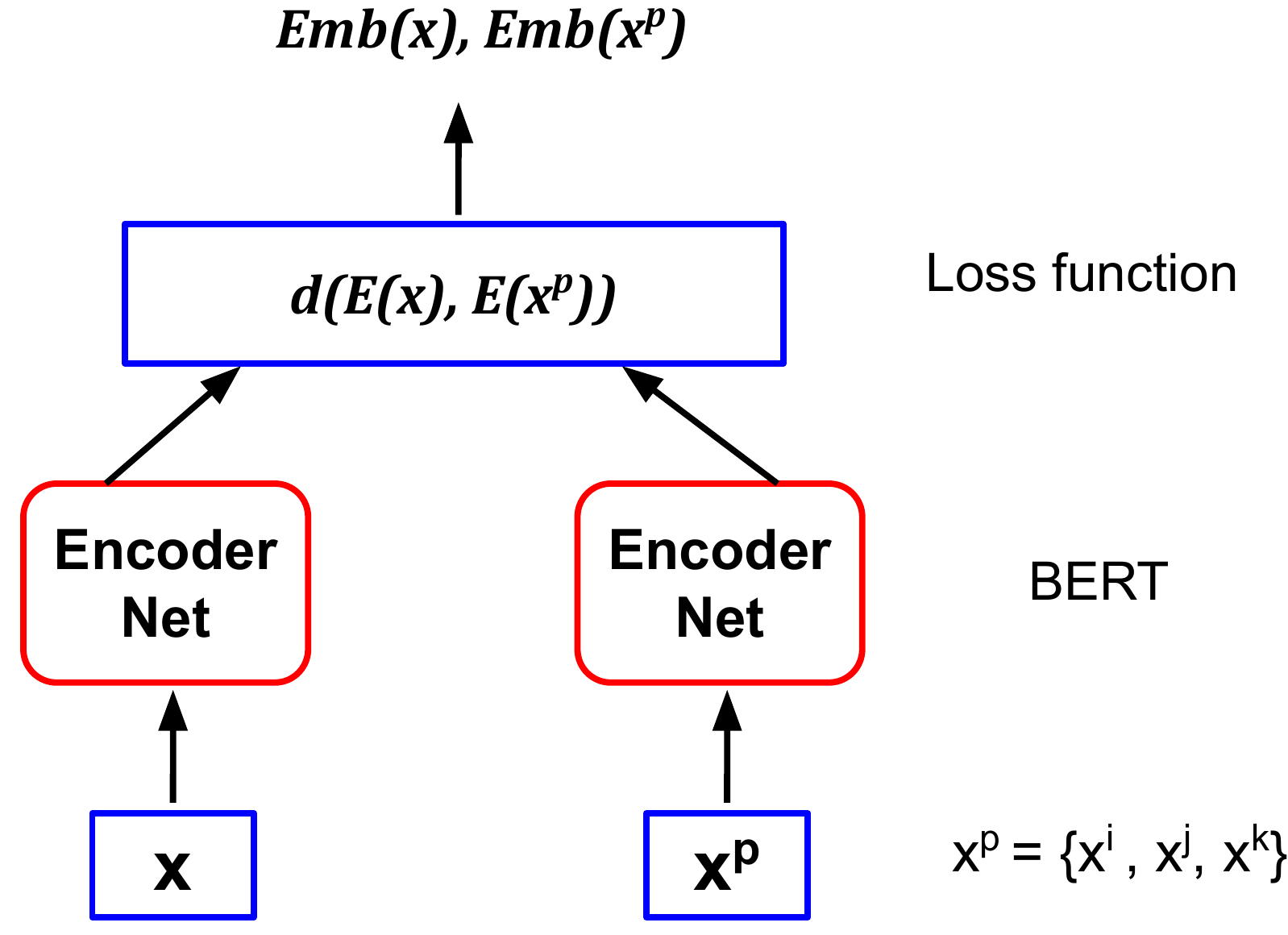}
  \caption{Discovering Novel Intent Categories}
  \label{fig:stage2}
\end{subfigure}
\caption{Overview of both stages of our approach, \mname{}}
\label{fig:advin}
\end{figure}

\smallskip
\noindent \textbf{Step I:} 
Prior open classification literature has modeled the problem of finding \textit{novel} or \textit{unseen} classes as an $(S+1)$-class  classification problem, with $S$ \textit{seen} classes and an additional \textit{unseen} class. Labeled training data is available for the seen classes, and an input instance is classed as unseen if it does not belong to any of the seen classes~\cite{xu2019open,shu2018unseen,shu2017doc}. 
In a similar vein, to classify utterances as containing a novel intent or not, we learn a BERT based multi-class classifier (Figure~\ref{fig:stage1}). 
Given an input utterance $\bm{x}$, it is converted into a sequence of tokens to be input to BERT. A special classification embedding \texttt{[CLS]} is added as the first token. 
The output of BERT is an $n$-dimensional utterance encoding $(\bm{e_1},...,\bm{e_n})$. A multi-class intent classification layer on top of BERT predicts the utterance intent based on the final hidden state $\bm{e_1}$ of the \texttt{[CLS]} token as: 

\begin{center}
$y = \textrm{softmax}(\bm{W e_1} + \bm{b})$
\end{center}

\noindent where $\bm{W}$ is a task-specific parameter weight matrix. 
We fine tune the parameters $\theta$ in different BERT layers with different learning rates as in~\cite{howard2018universal}, 
as follows:

\begin{center}
$\theta_t^l = \theta_{t-1}^l - \eta^l \nabla_{\theta_l} J(\theta)$
\end{center}

\noindent where $\eta_l$ and $\theta_l$ represent the learning rate and parameters respectively of the $l$-th BERT layer. $\nabla_{\theta_l} J(\theta)$ is the gradient with respect to the model's objective function. We set the base learning rate to $\eta^L$ and use $\eta^{l-1} = \xi.\eta^l$, where $\xi$ is a decay factor $\leq$ 1. 
We fine-tune all BERT parameters as well as $\bm{W}$ jointly by maximizing the log-probability of the correct intent label.


For our $(S+1)$-class classification model described above, labeled training data $\mathcal{D_T}$ is available for the $S$ seen intents. We use \textit{out-of-domain} (OOD) intent detection datasets distinct from both $\mathcal{D_T}$ and $\mathcal{D_C}$, as training data for the $(S+1)$-th `novel' intent class. This OOD data comes from out-of-domain intent-labeled publicly available datasets (e.g. SNIPS~\cite{coucke2018}, ATIS~\cite{dahl1994}). Since we use already annotated, publicly available intent labeled data as OOD data, we don't require extra human annotation effort here. Note that while training, \mname{} only requires the information that the OOD intents do not overlap with those in $\mathcal{D_T}$ or $\mathcal{D_C}$, and \textit{not} the actual intent labels of the OOD data. 
However, if the OOD data is annotated with $m$ intent classes, \mname{} can use this information by fragmenting the $(S+1)^\text{th}$ class into $m$ classes, forming an $(S+m)$-class classification layer. Using $m$ classes may provide a better representation of the utterances likely to contain unseen intents (as shown in Section~\ref{sec:evalution}). An utterance classified into any of the $m$ classes is flagged as having a novel intent. 

\smallskip
\textbf{Step II:} \mname{} employs an additional check as per the DOC algorithm~\cite{shu2017doc}. DOC learns statistical confidence thresholds for each seen intent class $s_i$. DOC thus captures instances that have been classified to one the $S$ seen intents with a low confidence. 
If the class-specific prediction probabilities for an input utterance are less than the thresholds learned for each seen intent $s_i$, that utterance is also classified as having a novel intent (see Figure~\ref{fig:stage1}).






\subsection{Stage II: Discovering Novel Intent Categories}
\label{sec:stage2}

The output of the previous step of \mname{} gives us all utterances $\mathcal{D_X}$ that potentially contain a newly emerging intent. The next step is to discover the actual latent intent categories $U$ within $X$. We use complete-linkage, agglomerative hierarchical clustering~\cite{gowda1978agglomerative} to group together related utterances in $\mathcal{D_X}$, and find the potential novel intents $U$. 

\smallskip

\noindent \textbf{Knowledge Transfer Component: }
Intuitively, humans seem to understand and categorize newly encountered objects based on the characteristics of similarities or differences that they learn from their prior knowledge of similar or comparable objects. We utilize a similar idea in order to learn the distance threshold $\delta$ for hierarchical clustering. That is, we transfer the knowledge learned from clustering the utterances containing seen intents $S$, to the utterances containing novel intents.
First, we perform hierarchical clustering on the labeled training data utterances $\mathcal{D_T}$, using the seen intents as ground truth cluster labels. 
We obtain a distance value that maximizes the inter-intent distance between utterance clusters and minimizes the intra-intent distance within a cluster, by maximizing the F1 score of their clustering arrangement. 
In an ideal scenario, every obtained cluster $L_i$ represents a single seen intent. We also assume that utterances in the training set $\mathcal{D_T}$ of seen intents, as well as those in the unlabeled corpus $\mathcal{D_X}$ of newly emerging unseen intents come from similar distributions. 
Our final step is then to \textit{transfer} this distance threshold $\delta$ learnt from the \textit{seen} intent utterances, to hierarchically cluster utterances in $\mathcal{D_X}$ containing novel or unseen intents. The distance threshold $\delta$ is defined as $\max\limits_{x \in L_i, x^p \in L_j} f(Emb(\bm{x}), Emb(\bm{x^p}))$ according to complete-linkage hierarchical clustering. Here, function $f(.)$ quantifies the distance between the embeddings $Emb(.)$ of an utterance pair $(\bm{x}$, $\bm{x^p})$ belonging to clusters $L_i$ and $L_j$ respectively.  





We now describe how we learn the distance function $f$ between pairs of utterances.

\smallskip

\noindent \textbf{Learning Pairwise Utterance Distances for Clustering: }
We learn a neural network model to obtain utterance embeddings $Emb(.)$ for both the labeled corpus of seen intents ($\mathcal{D_T}$) and the unlabeled utterances detected to have novel intents ($\mathcal{D_X}$). 
To train this model (see Figure~\ref{fig:stage2}), we create a training dataset from $\mathcal{D_T}$ comprising pairs of utterances $(x, x^p)$. Both $x$ and $x^p$ contain a seen intent. For each $x$, there are three possible choices for its paired utterance $x^p$: (i) $x^i$ containing the same domain and intent as $x$, (ii) $x^j$ containing the same domain but different intent than $x$, and (iii) $x^k$ containing a different domain and intent than $x$. Thus, $x^p \in \{x^i, x^j, x^k\}$. 
These utterance pairs $(x, x^p)$ are fed as input to an \texttt{EncoderNet}, which consists of a BERT transformer block. We use the same learning rate decay strategy of fine-tuning the BERT layers as earlier. Representations $E(\bm{x})$ and $E(\bm{x^p})$ are learned by the second last BERT layer (Figure~\ref{fig:stage2}).
The next layer on top of the \texttt{EncoderNet} blocks uses a distance function $d$ to compute pairwise representation distances, subject to the following bi-directional constraints: (i) the distance $d(E(x), E(x^i))$ between the representations of $x$ and $x^i$ should be less than $d(E(x), E(x^j))$; (ii) $d(E(x), E(x^i))$ between the representations of $x$ and $x^i$ should be less than $d(E(x), E(x^k))$; and (iii) the distance $d(E(x), E(x^j))$ between the representations of $x$ and $x^j$ should be less than the distance $d(E(x), E(x^k))$ between the representations of $x$ and $x^k$. These constraints utilize the semantic relationships between the utterance pairs containing different types of seen domains and seen intents. 
We then formulate a loss function $\mathfrak{L}$ to train our model, given by:


\begin{align*}
& \frac{1}{M} \sum\limits_{i, j, k} \{\max[0, m_1 + d(E(x), E(x^i)) - d(E(x), E(x^j)] + \alpha \max[0, m_2 + d(E(x), E(x^i)) - d(E(x), \\
& E(x^k)] + \beta \max[0, m_3 + d(E(x), E(x^j)) - d(E(x), E(x^k)] \}    
\end{align*}

\noindent where $m_1$, $m_2$, $m_3$ are predefined margins, $\alpha$ and $\beta$ are predefined weighting scalars and $M$ is the total number of utterance pairs $(x, x^p)$. We found such a loss formulation to outperform the popular contrastive loss~\cite{hadsell2006dimensionality} and triplet loss~\cite{schroff2015facenet} functions (Tables~\ref{tab:stage21} and~\ref{tab:stage22}). 
Next, a non-linear activation followed by a linear layer outputs embeddings $Emb(\bm{x})$ and $Emb(\bm{x^p})$ for the pair $(x, x^p)$. 
Finally, pairwise distances $f(Emb(\bm{x}), Emb(\bm{x_p}))$ are computed between all utterance pairs in this embedding space, to be given as input to 
hierarchical clustering. 
Note that we do not require any intent or domain labels for the unlabeled corpus $\mathcal{D_X}$ containing the unseen (novel) intents.




\subsection{Stage III: Linking Related Novel Intents into Novel Domains}
\label{sec:stage3}
After hierarchical clustering, \mname{} discovers a set of 
novel emerging intents. 
We now aim to link together mutually related intents sharing the same broad functionality or the same semantic category into \textit{domains}. 
Hierarchical clustering already gives us a provision to `merge' together intents at the upper levels of the hierarchy based on distance. 
However, we find that using the domain labels available for the seen intents in a more direct manner leads to a better grouping of related novel intents into novel domains. We perform the following steps to link intents into domains, to create an intent-domain taxonomy:

(i) We assume that we have information regarding the domain of each seen intent in $S$. Assuming an ideal clustering in Section~\ref{sec:stage2}, each seen intent cluster $L_i$ will contain utterances belonging to a single seen intent $s_i \in S$. The domain label of cluster $L_i$ would be the domain of intent $s_i$ itself. However, $L_i$ may not be completely pure, i.e. it contains utterances with different seen intent labels. In such cases, we assign the domain label for $L_i$ as the domain of the intent of the majority of the utterances in $L_i$. 

(ii) We next obtain a representation $Emb_L(L_i)$ for each seen and novel intent cluster $L_i$, as the average of the embeddings $Emb(\bm{x})$ (from Figure~\ref{fig:stage2}) of all utterances $x \in L_i$. 

(iii) Finally, we re-use our Knowledge Transfer component to cluster the representations $Emb_L(L_i)$ of the seen intent clusters $L_i$ themselves. This time we use the seen \textit{domains} as ground truth cluster labels (instead of the seen \textit{intents} as in Section~\ref{sec:stage2}). As earlier, we obtain a distance threshold $\delta$ that maximizes the F1 score with respect to the seen domains. We then transfer this threshold to perform hierarchical clustering of the novel (unseen) intent clusters. 

\smallskip

\noindent Each cluster so obtained contains groups of related novel intents, representing novel domains. 
Thus, \mname{} creates a taxonomy of novel intents and domains from unlabeled user utterances.


\section{Evaluation}
\label{sec:evalution}

\subsection{Datasets and Experimental Setup}

We test \mname{} on the real-world datasets 
of SNIPS~\cite{coucke2018}
, ATIS~\cite{dahl1994}
, Facebook's task-oriented semantic parsing (FTOP) data~\cite{gupta2018semantic}, and 
Internal NLU Data\footnotemark from a commercial voice assistant. 
For evaluation on SNIPS, ATIS and the Internal data, we completely remove all utterances associated with certain random sets of intents and domains from the training and validation sets, detailed in Table~\ref{tab:removed}. The FTOP dataset has intent types labeled as `unsupported', so we simply remove all utterances belonging to these intents while training \mname{}. Treating these removed intent and domain categories as novel (unseen), we then assess the efficacy of \mname{} in discovering these intents and domains during the testing phase. 
ATIS dataset intents are about airline reservations and are relatively similar to each other, while the SNIPS dataset consists of relatively dissimilar intent types. 
Our empirical configurations therefore holistically exhibit the performance of \mname{} when the novel intents being discovered have varying degrees of similarity with each other (or with the existing `seen' intents).

\smallskip

\noindent \textbf{Hyperparameters:} 
We used the English uncased BERT-Base model~\cite{devlin2018bert} in all steps of \mname{}. It has $12$ transformer layers, $768$ hidden states, and $12$ self-attention heads. We kept the dropout probability at $0.1$ and used the Adam optimizer~\cite{kingma2014adam} with parameters $\beta_1= 0.9$ and $\beta_2= 0.999$. We used slanted triangular learning rates~\cite{howard2018universal} for BERT with the base learning rate at $2e^{-5}$, and warm-up proportion at $0.1$. We empirically set the batch size to $64$, and the number of training epochs to $8$ for detecting instances with novel intents, and $15$ for discovering the latent intent categories. 
To learn pairwise distances between utterances for clustering, we set $\alpha$ = $\beta$ = 1, and $m_1$ = $m_2$ = $m_3$ = 0.05. While discovering novel intent categories (Section~\ref{sec:stage2}), we used cosine similarity as the function `\textit{d(.)}' while computing $d(E(\bm{x}), E(\bm{x^p}))$, and euclidean distance as the function `\textit{f(.)}' while computing $f(Emb(\bm{x}), Emb(\bm{x^p}))$.

\begin{table}[t]
\caption{Evaluating \mname{} on discovering novel intents and domains removed during training}
\label{tab:removed} \centering
\small
\hskip-0.4cm\begin{tabularx}{1.03\linewidth}{L{0.15}|L{1.4}|L{0.17}|L{0.13}|L{0.15}}
\hline
\textbf{Dataset} & \textbf{Sets of intents removed from training data for evaluation} & \textbf{\# of data samples} & \textbf{Vocab size} & \textbf{Avg. text len.}  \\
\hline
SNIPS & \textbf{Set 1}: \textit{Weather, Restaurant}; \textbf{Set 2}: \textit{AddToPlaylist, RateBook} & 13.8K & 10.9K & 9.05 \\
ATIS & \textbf{Set 1}: \textit{airline, meal, airfare, day-name, distance}; \textbf{Set 2}: \textit{ flight-time, flight-no, flight, aircraft, ground-service}  & 5.87K & 0.87K & 11.2  \\
FTOP & \textbf{Set 1}:  \textit{unsupported, unsupported-event, unsupported-navigation, unintelligible} & 44.78K & 16.69K  & 8.93  \\
Internal NLU Dataset & \textbf{Set 1}: \textit{Weather, Calendar, Todos}; \textbf{Set 2}: \textit{Bookings\&Reservations, Sports, Local Search, Video, General Media}; \textbf{Set 3}: \textit{Recipe, Music, Shopping, Communication}; \textbf{Set 4}: \textit{Global, Knowledge}  & 3.16M & 26.7K & 3.72 \\
\hline
\end{tabularx}
\end{table}

\footnotetext{Customer utterances from a voice-powered virtual assistant, whose name we omit for the double-blind review purpose.}

\subsection{Baselines and Evaluation Metrics} 


We designate the first step of our proposed approach of detecting utterances with novel intents as \textbf{``\mname{} \textit{(m-unseen + DOC)}"}, and compare it with the following state-of-the-art approaches in Table~\ref{tab:stage1}: 

\smallskip
\noindent \noindent (i) \textbf{DOC}~\cite{shu2017doc}: uses a CNN with a 1-vs-rest sigmoid layer on top. It tightens the sigmoid decision boundary by learning class-specific confidence thresholds to detect novel intents. 

\noindent (ii) \textbf{IntentCapsNet}~\cite{xia2018}: uses capsule neural networks in a zero shot setting to discover newly emerging intents. 

\noindent (iii) \textbf{LOF-LMCL}~\cite{lin2019deep}: uses local outlier detection on top of a Bi-LSTM trained with a large margin cosine loss to classify seen and unseen intents.

\noindent (iv) \textbf{\mname{} \textit{(1-unseen)}/\mname{} \textit{(1-unseen + DOC)}}: variants of \mname{} using an \textit{(S+1)}-class classification model (Section~\ref{sec:stage1}), with and without the additional check 
per the DOC heuristic. 

\noindent (v) \textbf{\mname{} \textit{(m-unseen)}}: \mname{}'s \textit{(S+1)}-th class split into $m$ novel classes, without the DOC heuristic. 

\smallskip

\begin{table*}[t]
\centering
\caption{F1-score of various approaches for detecting if an utterance contains a novel intent or not.} 
\begin{tabularx}{\linewidth}{L{0.87}|L{0.11}|L{0.11}|L{0.11}|L{0.11}|L{0.25}|L{0.11}|L{0.11}|L{0.11}|L{0.11}} \hline
  \textbf{Approach} & \multicolumn{2}{c|}{\textbf{SNIPS}} & \multicolumn{2}{c|}{\textbf{ATIS}} & \textbf{FTOP} & \multicolumn{4}{|c}{\textbf{Internal Dataset}}  \\\cline{2-10}
    &  \textbf{Set1}   & \textbf{Set2}  &  \textbf{Set1}  & \textbf{Set2}  &  \textbf{Set1}  &  \textbf{Set1}  & \textbf{Set2} & \textbf{Set3}  & \textbf{Set4}  \\
    \hline
  DOC~\cite{shu2017doc}  & 0.73 & 0.69 & 0.71 & 0.7 & 0.76 & 0.7 & 0.73 & 0.72 & 0.71 \\ 
  IntentCapsNet~\cite{xia2018} & 0.81 & 0.77 & 0.7 & 0.75 & 0.8 & 0.82 & 0.83 & 0.78 & 0.8 \\ 
  LOF-LMCL~\cite{lin2019deep} & 0.79 & 0.73 & 0.68 & 0.74 & 0.78 & 0.84 & 0.8 & 0.82 & 0.81\\ 
  \mname{} \textit{(1-unseen)} & 0.76 & 0.73 & 0.68 & 0.72 & 0.78 & 0.77 & 0.75 & 0.79 & 0.8 \\ 
  \mname{} \textit{(1-unseen+DOC)} & 0.85 & 0.81 & 0.73 & 0.8 & 0.87 & 0.86 & 0.87 & 0.88 & 0.86\\ 
  \mname{} \textit{(m-unseen)} & 0.78 & 0.75 & 0.7 & 0.75 & 0.8 & 0.8 & 0.8 & 0.82 & 0.84 \\ 
  \mname{} \textit{(m-unseen+DOC)} & \textbf{0.9} & \textbf{0.87} & \textbf{0.78} & \textbf{0.84} & \textbf{0.9} & \textbf{0.9} & \textbf{0.92} & \textbf{0.9} & \textbf{0.9} \\ 
\hline
\end{tabularx}
\label{tab:stage1}
\end{table*}

\begin{table*}[t]
\centering
\caption{Discovering the latent intent types for utterances with novel intents. `\#int.' shows the number of discovered intents, `GT' denotes the true number of intents, and `Pur.' denotes cluster purity.}\smallskip
\begin{tabularx}{1.02\linewidth}{L{0.66}|L{0.12}|L{0.11}|L{0.12}|L{0.09}|L{0.12}|L{0.11}|L{0.12}|L{0.09}|L{0.12}|L{0.11}|L{0.12}|L{0.08}} \hline
  \textbf{Approach} & \multicolumn{4}{c|}{\textbf{SNIPS Set 1 (GT = 2)}} & \multicolumn{4}{c|}{\textbf{SNIPS Set 2 (GT = 2)}} & \multicolumn{4}{|c}{\textbf{FTOP Set 1 (GT = 4)}}  \\\cline{2-13}
    &  \footnotesize{\bf \#int.} & \textbf{NMI}  &  \textbf{Pur.}   &  \textbf{F1} &  \footnotesize{\bf \#int.} & \textbf{NMI}  &  \textbf{Pur.} &  \textbf{F1} &  \footnotesize{\bf \#int.} & \textbf{NMI}  &  \textbf{Pur.}   &  \textbf{F1}
    \\
    \hline
  \mname{} \textit{(clf+hier)} & \textbf{3} & 0.78 & 0.9 & 0.76 & \textbf{3} & 0.7 & 0.8 & 0.69 & 24 & 0.4 & 0.56 & 0.38\\ 
  \mname{} \textit{(triplet+hier)} & 4 & 0.71 & 0.81 & 0.69 & 5 & 0.65 & 0.76 & 0.66 & 48 & 0.36 & 0.51 & 0.35\\ 
  \mname{} \textit{(ProdLDA)} & NA & 0.71 & 0.84 & 0.72 & NA & 0.66 & 0.79 & 0.68 & NA & 0.42 & 0.53 & 0.38\\ 
  \mname{} \textit{(pair+RCC)} & \textbf{3} & 0.79 & 0.9 & 0.77 & \textbf{3} & 0.7 & 0.8 & 0.7 & 30 & \textbf{0.5} & \textbf{0.63} & 0.41\\ 
  \mname{} \textit{(pair+hier)} & \textbf{3} & \textbf{0.8} & \textbf{0.92} & \textbf{0.78} & \textbf{3} & \textbf{0.72} & \textbf{0.83} & \textbf{0.71} & \textbf{19} & 0.46 & 0.61 & \textbf{0.51}\\ 
\hline
\end{tabularx}
\label{tab:stage21}
\end{table*}

\begin{table*}[t]
\centering
\caption{Discovering the actual, latent novel intents and novel domains of input utterances. For Internal Data Sets 1 and 2, the first two columns show the number of new intents (\#int.) and new domains (\#dom.) discovered respectively. `GT ($d$, $i$)' denotes the true number of domains $d$ and intents $i$.}\smallskip
\begin{tabularx}{\linewidth}{L{0.65}|L{0.15}|L{0.17}|L{0.11}|L{0.14}|L{0.1}|L{0.16}|L{0.17}|L{0.11}|L{0.14}|L{0.1}} \hline
  \textbf{Approach} & \multicolumn{5}{c|}{\textbf{Internal Data Set 1 (GT = 3, 22)}} & \multicolumn{5}{|c}{\textbf{Internal Data Set 2 (GT = 5, 53)}}  \\\cline{2-11}
    &  \footnotesize{\bf \# int.} & \footnotesize{\bf \# dom.} & \textbf{NMI}  &  \textbf{Purity}   &  \textbf{F1} &  \footnotesize{\bf \# int.} & \footnotesize{\bf \# dom.} & \textbf{NMI}  &  \textbf{Purity} &  \textbf{F1}
    \\
    \hline
  \mname{} \textit{(clf+hier)} & 108 & 35 & 0.53 & 0.69 & 0.48 & 178 & 71 & 0.41 & 0.64 & 0.36  \\ 
  \mname{} \textit{(triplet+hier)} & 206 & 51 & 0.52 & 0.63 & 0.41 & 205 & 65 & 0.33 & 0.61 & 0.31  \\ 
  \mname{} \textit{(ProdLDA)} & NA & NA & 0.56 & 0.7 & 0.55 & NA & NA & 0.4 & 0.69 & 0.42   \\ 
  \mname{} \textit{(pair+RCC)} & 91 & 40 & 0.55 & 0.73 & 0.52 & 188  & 57 & 0.5 & 0.74 & 0.5 \\ 
  \mname{} \textit{(pair+hier)} & \textbf{74} & \textbf{29} & \textbf{0.6} & \textbf{0.75} & \textbf{0.6} & \textbf{167} & \textbf{36}  & \textbf{0.57} & \textbf{0.75} & \textbf{0.55} \\ 
\hline
\end{tabularx}
\label{tab:stage22}
\end{table*}


The first stage of discovering utterances with novel intents (Section~\ref{sec:stage1}) is evaluated using the standard F1-score metric. For evaluating the next two stages of discovering the actual intent categories in the user utterances and linking the newly discovered intents into domains (Sections~\ref{sec:stage2} and~\ref{sec:stage3}), we designate our proposed method as \textbf{``\mname{} \textit{(pair+hier})"}, i.e. using distances between embeddings learned via our pairwise margin loss function as input to hierarchical clustering. 
As per our knowledge, work in the literature only detects if text utterances contain novel intents or not, and does not identify the actual, latent intent categories in unlabeled text. Therefore, we compare \mname{} with its own variant baselines: 

\smallskip

\noindent (i) \textbf{\mname{} \textit{(clf+hier)}}: uses the representation learned by the 2nd to last BERT layer of our classification model in Figure~\ref{fig:stage1}, as input to hierarchical clustering. 

\noindent (ii) \textbf{\mname{} \textit{(triplet+hier)}}: uses embeddings learned by a triplet network~\cite{schroff2015facenet} as input to hierarchical clustering. The inputs to the network are utterance triplets $(x, x^-, x^+)$. $x^-$ and $x^+$ contain the same domain and intent, and different domain and intent as utterance $x$ respectively. 

\noindent (iii)  \textbf{\mname{} \textit{(ProdLDA)}}: uses a neural topic modeling method ProdLDA~\cite{srivastava2017autoencoding} to discover novel intent categories, instead of clustering. 
ProdLDA requires the number of topics as input, so we give it the number of clusters output by ``\mname{} \textit{(pair+hier)}".

\noindent (iv) \textbf{\mname{} \textit{(pair+RCC)}}: uses embeddings learned via our pairwise loss function as input to the Robust Continuous Clustering algorithm (RCC)~\cite{shah2017robust}, instead of hierarchical clustering. 
\smallskip

Ground truth intent and domain labels are available for the various experimental data Sets we used for evaluation (Table~\ref{tab:removed}). We thus use the following standard clustering metrics to evaluate the novel intents and domains discovered: (i) Comparing the number of discovered intents and domains to the ground truth number. (ii) Normalized Mutual Information (NMI): a normalization of the mutual information by a generalized mean of the entropy of the ground truth and the entropy of the predicted cluster labels. 
(iii) Purity: the extent to which a cluster contains utterances with a single intent or domain. (iv) F1 score.

\subsection{Results}

\noindent \textbf{Baseline Comparison: } Table~\ref{tab:stage1} shows the F1-score of various approaches, on classifying an intent as novel or not on different datasets with different dataset configurations. We observe that our approach using an $(S+m)$-class classifier during training and the DOC heuristic, ``\mname{} \textit{(m-unseen+DOC)}", outperforms all baselines on all datasets by at least 6\% F1 score points. \mname{} also outperforms the zero shot IntentCapsNet model (that uses relevant information available beforehand about the new test intents to be discovered), without using any prior knowledge about the novel intents or domains.

Tables~\ref{tab:stage21} and~\ref{tab:stage22} show the performance of different approaches in discovering intent categories for the utterances containing novel intents as predicted in Table~\ref{tab:stage1}. We observe that our proposed approach ``(\mname{} + \textit{pair+hier})" using distances learned via our pairwise loss training method along with hierarchical clustering outperforms the rest of the baselines. The learned intent-clusters have a purity value more than 80\% for the SNIPS dataset. Purity decreases to $61\%$ for FTOP, primarily because semantically diverse utterances have been given the same ground truth intent label of \textit{`unsupported'} or \textit{`unsupported-event'}. The `NA' value for ``\mname{} \textit{(ProdLDA)}" in the column `\# int.' denotes that the \textit{ProdLDA} algorithm by design does not output the number of new intents discovered. It takes this as input in the form of the number of topics. We observe similar empirical trends as above on ATIS Sets 1 and 2, and Internal NLU Data Sets 3 and 4. In the interests of space, we do not present these results.

\begin{figure}
\begin{minipage}[c][7cm][t]{.5\textwidth}
  \vspace*{\fill}
  \centering
  {\label{fig:tsne}\includegraphics[width=8.6cm,height=6cm]{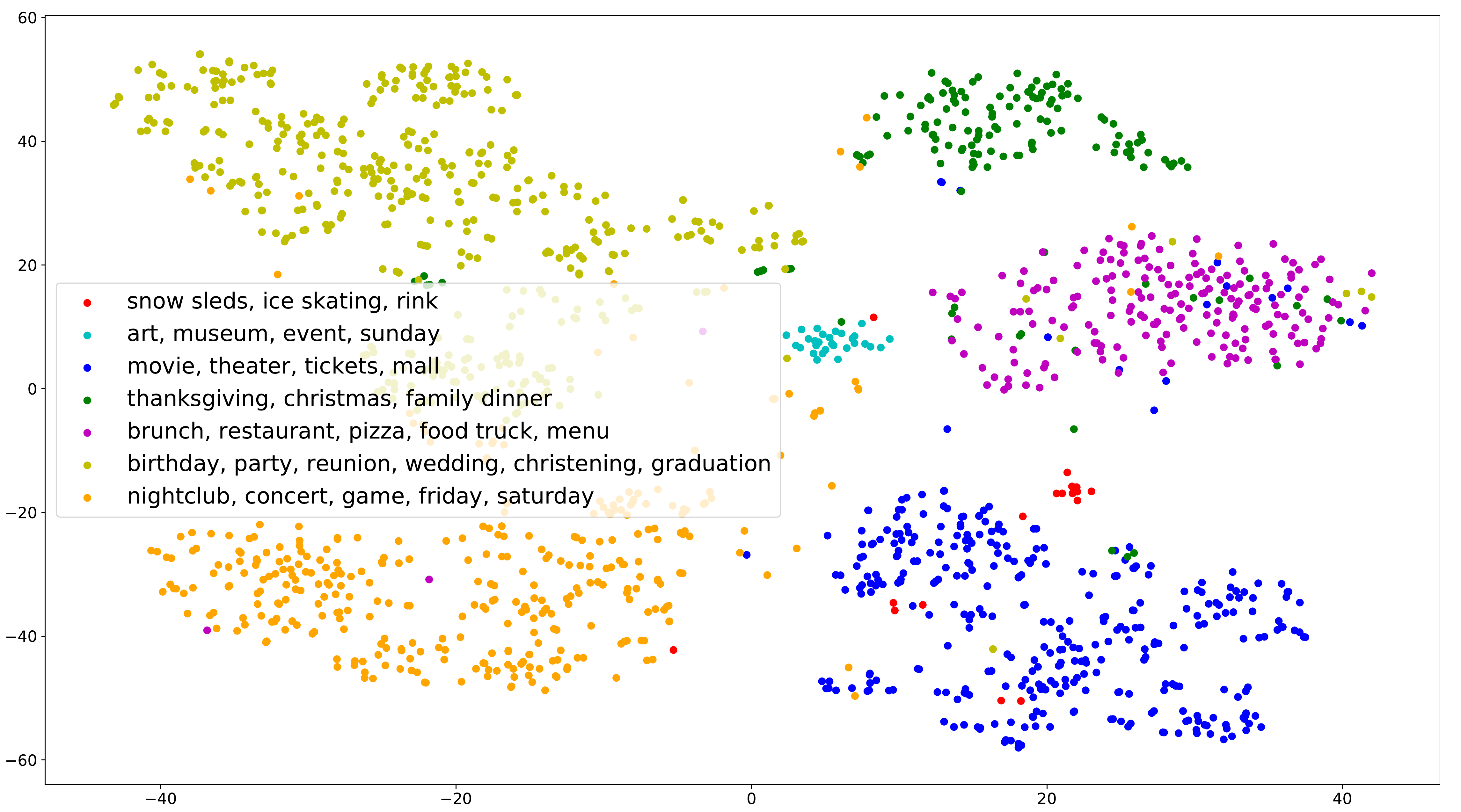}}
  \subcaption{}
\end{minipage}%
\begin{minipage}[c][7cm][t]{.5\textwidth}
  \vspace*{\fill}
  \centering
  {\label{fig:num_ood_intents}\includegraphics[width=6cm,height=2.8cm]{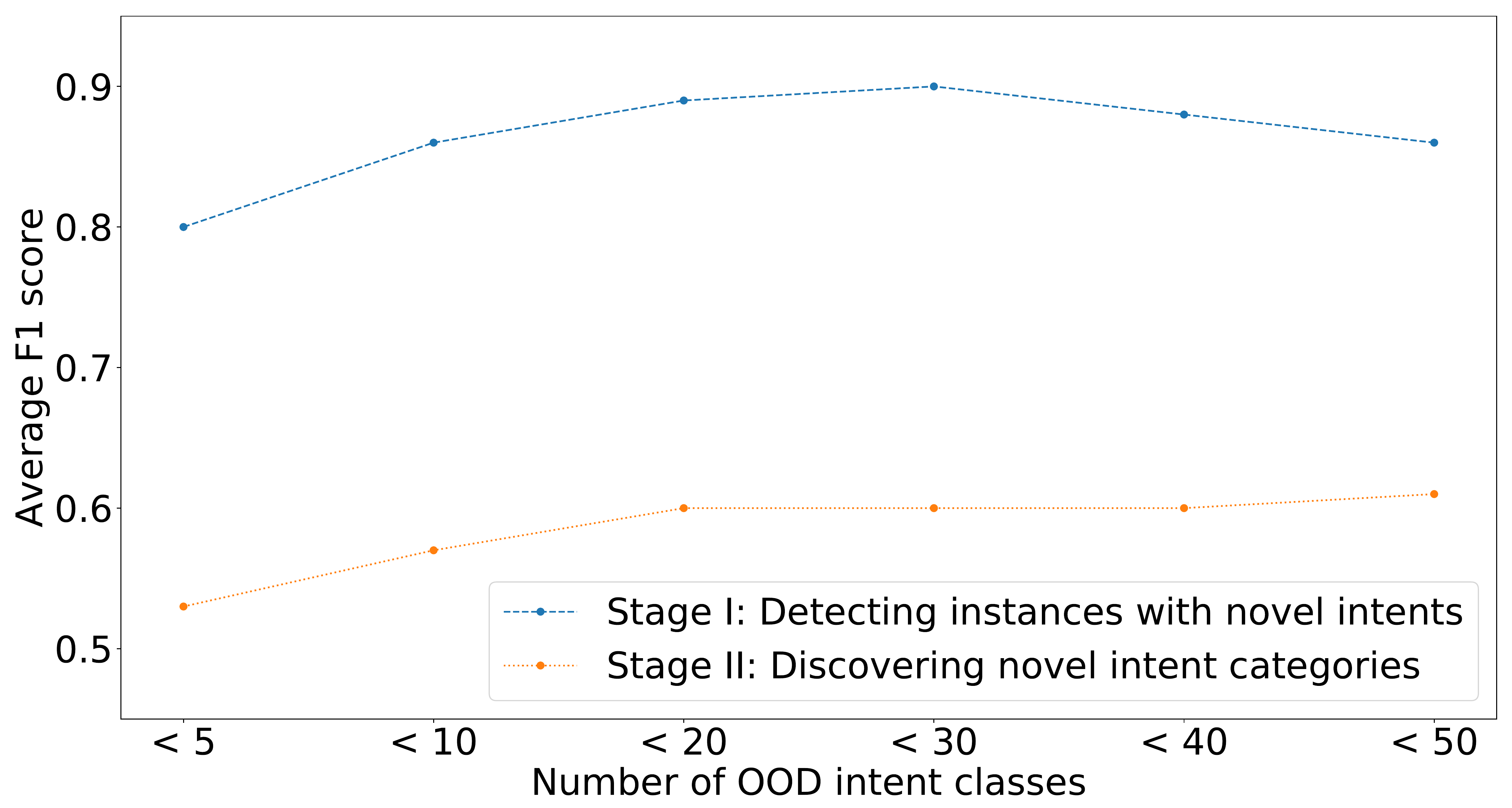}}
  \subcaption{}
  \par\vfill
  {\label{fig:num_ood_intents}\includegraphics[width=6cm,height=2.8cm]{num_ood_intents_f1}}
  \subcaption{}
\end{minipage}
\caption{(a) Visualizing embeddings learned by \mname{} for novel intents discovered within the \textit{`unsupported'} FTOP category. The figure also shows the varying F1 score averaged across (b) number of OOD intent classes and (c) number of labeled OOD instances, given as training data for Stage I.}
\end{figure}

Our technique discovers $1.5$-$4.5$ times more novel intents than present in the ground truth, as seen from the `\# int.' columns in Tables~\ref{tab:stage21} and~\ref{tab:stage22}. To investigate this further, we use t-SNE~\cite{maaten2008visualizing} to visualize the embeddings learned by training our proposed \mname{} method, for the utterances belonging to the \textit{`unsupported'} intent category of the FTOP dataset. In Figure~\ref{fig:tsne}, each color represents a different novel intent category discovered. We also show the most frequent utterance words present in each newly emergent intent. 
We observe that \textit{`Unsupported'} has been split up by \mname{} into $7$ finer-grained, semantically sensible novel intents. For instance, the utterances \textit{``What city has the most traffic in the US"} and \textit{``Family friendly bars near me"}, from the \textit{`unsupported'} intent category, have been separated by \mname{} into different intent categories.  Figure~\ref{fig:tsne} also explains the lower performance values in Table~\ref{tab:stage21} for the FTOP dataset. We thus find that \mname{} largely learns semantically appropriate, discriminative representations for the newly emerging intents (also echoed by our user study in Table~\ref{tab:user study}).

\begin{table}[t]
\centering
\caption{Evaluating \mname{}'s prediction of utterance pairs containing the same novel intent or not. Columns show F1 scores w.r.t  using (i) user study and (ii) dataset-provided annotations as ground truth.}\smallskip
\begin{tabularx}{1.03\linewidth}{L{0.45}|L{0.78}|L{0.77}} \hline
  \textbf{Approach} & \textbf{FTOP:F1 user study(F1 Dataset)} & \textbf{Internal:F1 user study(F1 Dataset)} \\ \hline
  \mname{} \textit{(clf+hier)} & 0.66 (0.6) & 0.6 (0.68) \\ 
  \mname{} \textit{(triplet+hier)} & 0.6 (0.57) & 0.55 (0.61)  \\ 
  \mname{} \textit{(ProdLDA)} & 0.65 (0.56) & 0.54 (0.61)  \\ 
  \mname{} \textit{(pair+RCC)} & 0.64 (0.6) & 0.56 (0.64)  \\ 
  \mname{} \textit{(pair+hier)} & 0.7 (0.61) & 0.61 (0.71)  \\ 
\hline
\end{tabularx}
\label{tab:user study}
\end{table}

\begin{table}[t]
\centering
\caption{\mname{} discovering new intent categories for unlabeled utterances with novel intents, using limited supervision as pairwise constraints. Parentheses show original results without using supervision.}\smallskip
\begin{tabularx}{\linewidth}{L{0.445}|L{0.345}|L{0.41}|L{0.41}|L{0.39}} \hline
  \textbf{Dataset} & \textbf{\# intents} & \textbf{NMI} & \textbf{Purity} & \textbf{F1 score}
    \\
    \hline
  SNIPS Set 1 & 2 (3) & 0.82 (0.8) & 0.94 (0.92) & 0.85 (0.78) \\ 
  ATIS Set 1 & 12 (13) & 0.74 (0.72) & 0.77 (0.73) & 0.75 (0.72)  \\ 
  FTOP Set 1 & 17 (19) & 0.5 (0.46) & 0.64 (0.61) & 0.54 (0.51)  \\ 
  Internal Data Set 1 & 45 (56) & 0.65 (0.61) & 0.75 (0.71)  &  0.65 (0.62) \\ 
  Internal Data Set 2 & 82 (92) & 0.58 (0.54) & 0.7 (0.69)  & 0.61 (0.58)  \\ 
\hline
\end{tabularx}
\label{tab:supervision}
\end{table}


\smallskip

\noindent \textbf{Effect of OOD training data: }
We analyze the effect of the labeled OOD instances given as training input to Stage I (detecting utterances with novel intents) of \mname{}. Figures~\ref{fig:num_ood_intents} and~\ref{fig:num_ood_data} show that the performances of both Stage I (Section~\ref{sec:stage1}) and II (Section~\ref{sec:stage2}) of \mname{} improve with increase in the number of (i) OOD intent classes, and (ii) OOD training data examples; up to a particular threshold `m' for the Internal NLU Data Set I. Increasing `m' beyond this value leads to a performance drop. We also observe that the presence of finer-grained labeled utterances (e.g. the intent labels of ATIS) in the OOD data has a higher performance impact, than the coarser-grained intent labels (e.g. FTOP intents). 
We find similar trends for the other datasets also, but do not present them due to space constraints. 

\smallskip

\noindent \textbf{Linking Intents to form Domains: }
There is no domain information available for the SNIPS, ATIS and FTOP data. Hence, we only evaluate the performance of linking intents into domains on the Internal dataset. The column of `\# dom.' in Table~\ref{tab:stage22} for Internal Data Sets 1 and 2 shows the number of newly discovered domains by \mname{} after linking mutually related newly discovered intents. 
As earlier, \mname{} obtains a finer-grained grouping of novel intents into novel domains. It finds $2$-$3$ times more number of domains than the dataset annotations. On further inspection, we found that the domains for the Internal datasets have been created by collating intents satisfying common business goals or customer needs, and are less geared towards semantics. Contrarily, \mname{} focuses on the semantic meaning of the utterances and performs a more fine-grained categorization of newly emerging intents into domains. This leads to a slight discord between the domains uncovered by \mname{}, and the domain annotations in the dataset. Such an over estimation of the number of novel intents or domains 
is often acceptable in a practical setting, since the granularity of domains and intents learned by \mname{} can be easily `coarsened' by merging together certain novel intents or domains, 
based on downstream requirements. One way to do this is by soliciting human feedback 
(see Table~\ref{tab:supervision}). 

\smallskip

\noindent \textbf{User Study: }
To compare the intent and/or domain labels provided with the datasets and human perception of novel intents and domains while evaluating \mname{}, we conduct a \textit{user study} on the FTOP and Internal datasets.  We recruit crowd workers on Amazon Mechanical Turk for the FTOP data and employees familiar with the Internal Dataset for this purpose. We provide a set of random utterance pairs predicted as having novel intents to the annotators, and ask them to indicate whether the pair is likely to belong to the same intent category or not. 
For both datasets we compute the F1-score in Table~\ref{tab:user study}, by comparing the output of \mname{} with that of the human annotators, for $2500$ FTOP utterance pairs (inter-annotator agreement Cohen's $\kappa = 0.78$) and $1100$ Internal dataset utterance pairs (Cohen's $\kappa = 0.9$). 
We observe that `\mname{} (\textit{pair+hier})' trained with the pairwise margin loss and hierarchical clustering significantly outperforms all baselines by at least 5\% on both datasets with respect to human evaluation.   

\smallskip

\noindent \textbf{Introducing Limited Supervision: }
There is often some partial supervision or background knowledge already known to humans or provided by domain experts, regarding the unlabeled user utterances. Utilizing this can enhance the quality of the novel intents and domains discovered by \mname{}. Therefore, we test \mname{} in a semi-supervised setting, instead of the unsupervised hierarchical clustering, where we provide limited prior knowledge in the form of two types of pairwise constraints: (i) \textit{must-link}, for the utterances that must belong to the same novel intent category, and (ii) \textit{cannot-link}, for the utterances that cannot contain the same novel intent. We incorporate the constraints during hierarchical clustering by modifying the learned distance values between the utterance pairs. We show in Table~\ref{tab:supervision} the performance of \mname{}, while randomly selecting $3$ groups of $4$ utterances each for the \textit{must-link} and \textit{cannot-link} constraints. Thus, supervision is provided for $<25$ utterances. We observe a 2-8\% gain 
across various metrics and datasets. 
This experiment demonstrates that \mname{} can be easily extended to a semi-supervised setting, and minimal supervision if available, can significantly improve the quality of the discovered intents and domains over an unsupervised setting. 

\section{Conclusion}

We propose a novel and flexible multi-stage framework, \mname{}, to discover newly emerging, unknown intents and domains in large volumes of unlabeled text data. We first identify all input utterances likely to contain a novel intent. We next develop a network that learns discriminative deep features by maximizing inter-intent variance and minimize intra-intent variance between utterance pairs. We then transfer knowledge learned from intents seen during training to the unlabeled data containing novel intents. Finally, we hierarchically link mutually related intents into domains, to obtain a taxonomy of newly discovered intents and domains. We extensively evaluate \mname{} on three public benchmark datasets and real user utterances from a commercial dialog agent, and achieve state-of-the-art results across various empirical configurations. In future, we plan to (i) extend our proposed framework to handle input utterances consisting of multiple intents and/or domains per utterance; and (ii) use additional knowledge to better model human-perceived latent intents and domains.


\begin{small}
\bibliographystyle{coling}
\bibliography{main.bib}
\fontsize{9.0pt}{10.0pt}
\selectfont
\end{small}

\end{document}